\documentclass[times, 10pt,twocolumn]{article} 
\usepackage{latex8}
\usepackage{times}
\usepackage{amsmath}
\usepackage{ amssymb }
\usepackage{textcomp}
\usepackage{gensymb}
\usepackage{mathrsfs}
\usepackage{graphicx}
\usepackage{caption}
\usepackage{subfig}
\usepackage[export]{adjustbox}
\usepackage{xcolor}
\usepackage{makecell}
\usepackage{authblk}
\def\I{\mathbf{I}}
\def\M{\mathbf{M}}
\def\Mh{\hat{\M}}

\makeatletter
\def\thanks#1{\protected@xdef\@thanks{\@thanks
        \protect\footnotetext{#1}}}
\makeatother

\begin{document}

\graphicspath{ {figures/} }

\author[1]{Emily R. Bartusiak}
\author[1]{Sri Kalyan Yarlagadda}
\author[1]{David G\"{u}era}
\author[1]{Fengqing M. Zhu}
\author[2]{Paolo Bestagini}
\author[2]{Stefano Tubaro}
\author[1]{Edward J. Delp
\thanks{This material is based on research sponsored by DARPA and Air Force Research Laboratory (AFRL) under agreement number FA8750-16-2-0173. The U.S. Government is authorized to reproduce and distribute reprints for Governmental purposes notwithstanding any copyright notation thereon. The views and conclusions contained herein are those of the authors and should not be interpreted as necessarily representing the official policies or endorsements, either expressed or implied, of DARPA and Air Force Research Laboratory (AFRL) or the U.S. Government.}}

\affil[1]{\normalsize Department of Electrical and Computer Engineering,  Purdue University, West Lafayette}
\affil[2]{Department of Electronics, Information and Bioengineering,   Politecnico di Milano, Italy\vspace*{-3em}}

\title{Splicing Detection And Localization In \\Satellite Imagery Using Conditional GANs \vspace*{-1em}}

\maketitle

\thispagestyle{empty}

\begin{abstract}
The widespread availability of image editing tools and improvements in image processing techniques allow  image manipulation to be very easy.  
Oftentimes, easy-to-use yet sophisticated image manipulation tools yields distortions/changes imperceptible to the human observer.  
Distribution of forged images can have drastic ramifications, especially when coupled with the speed and vastness of the Internet.  
Therefore, verifying image integrity poses an immense and important challenge to the digital forensic community.  Satellite images specifically can be modified in a number of ways, including the insertion of objects to hide existing scenes and structures.  
In this paper, we describe the use of a Conditional Generative Adversarial  Network (cGAN) to identify the presence of such spliced forgeries within satellite images.  Additionally, we identify their locations and shapes.  
Trained on pristine and falsified images, our method achieves high success on these detection and localization objectives.  
\end{abstract}

\section{Introduction}

Proper communication at both the public and personal level is key to the healthy development of human civilization. 
Over the years the means of communication have evolved, and in the present day the Internet is the most popular and important platform for communication. 
Many social media systems have developed using the Internet, and they provide a very cheap and effective way to express and shares one's ideas with the rest of the world. 
While an effective communication system for sharing information could help us become more informed and connected as a society, it could also be used to spread misinformation to achieve a nefarious objective. Hence, it is of paramount importance that we verify and authenticate the shared data on these systems.

While there are many ways of communicating ideas, such as speech, symbols,  and written text, images are today one of the most popular means. 
Unfortunately, manipulating images has become very easy. 
Tools such as GIMP and Photoshop can be used to manipulate images in a wide variety of ways, and they are easily accessible to the general public. 
To address this problem, the forensic community has developed a wide variety of tools to detect various kinds of image forgeries \cite{Rocha2011, Piva2013, Stamm2013}. 
While most of the images shared on the internet come from consumer cameras and smart-phones, other types of imagery such as satellite images are also very important in business and government applications and thus pose new problems for the forensic community \cite{bbc, russia}. 

With the increase in the number of satellites equipped with imaging sensors and the technological advancements made in satellite imaging technology, high resolution images of the ground are becoming popular. It is now possible to not only access these overhead images from public websites \cite{freesat} but also to buy custom satellite imagery of specific locations. Just like any other image, satellite images can also be doctored. While the forensic community has been developing tools to address forgeries of all types, they have been biased towards imagery captured from consumer cameras and smartphones \cite{Barni2010, Kirchner2015, Cozzolino2015, Bondi2017a}. The nature of acquisition of satellite imagery is quite different from that of images from consumer cameras hence the importance that forensic tools be developed that specifically target satellite imagery.

In the recent years, some methods \cite{Ho2005, Sri-ei, Ali2017} for satellite image forgeries have been developed. In \cite{Ho2005} Ho et al. have proposed an active forensic method based on watermarks to verify the authenticity of a satellite image. While watermarks are an effective way of ascertaining whether an image is forged or not, their absence renders such methods ineffective. In \cite{Ali2017} Ali et al. have proposed a passive  method based on machine learning to detect inpainting in satellite images. Yarlagadda et al. \cite{Sri-ei} have proposed a method based on deep learning to detect splicing in satellite images. They employ Generative Adversarial Networks (GANs)  \cite{GANIan, Goodfellow2016} to learn a compact representation of pristine satellite images and use it to detect splicing of various sizes.

In this paper, we discuss the detection and localization of splicing in satellite images. Splicing refers to replacement of pixels of a region of the image to add or remove an object. We employ a Conditional Generative Adversarial Network (cGAN) to  learn a mapping from a satellite image to its splicing mask. The trained cGAN operates on a satellite image of interest and outputs a mask of the same resolution that is indicative of the likelihood of a pixel belonging to a spliced region. Our cGAN's architecture is an extension of the popular pix2pix \cite{pix2pix}. 
Differently from \cite{Sri-ei}, we learn a direct mapping from an image to its forgery mask, rather than operating in a one-class fashion. To achieve this, we provide both pristine and spliced images to train our model, while the authors in \cite{Sri-ei} only use pristine images for training as they are trying to learn a compact representation of the pristine data and use it to identify forgeries.

We use the dataset proposed in \cite{Sri-ei} to validate our method. We report both the localization and detection performance.

\section{Problem Formulation}
We investigate the following two specific objectives in this paper: forgery detection and localization. $\textit{Detection}$ refers to the goal of determining if an RGB satellite image $\I$ has been modified via splicing. It is a binary classification problem where images can be considered $\textit{forged}$, if they have been modified, or $\textit{pristine}$, if not. $\textit{Localization}$ refers to the image segmentation goal of identifying each pixel in a forged image that belongs to the spliced entity, otherwise known as the $\textit{forgery}$.  These goals are defined in a similar manner to those outlined in \cite{Sri-ei}.  

Forgery masks $\M$ are used to help us visualize and determine the outcomes for these objectives. For an image $\I$, a forgery mask $\M$ of the same dimensions shows the forgery in $\I$ (if it exists).  In other words, for a satellite image $\I(x,y)$ where $(x,y)$ specifies the coordinate location of a pixel in $\I$, the corresponding forgery mask $\M$ is comprised of values defined as
\begin{equation} \label{eq1}
\M(x,y) = 
\begin{cases} 
      255 & \text{if $\I(x,y)$ is forged}, \\
      0 & \text{otherwise}.
   \end{cases}
\end{equation}
Therefore, the shape, size, and location of a forgery in an image $\I$ can be ascertained from the mask $\M$ if it contains white pixel values (i.e., 255). At an extreme, an entirely white mask $(\M \neq 0)$ indicates that every pixel in $\I$ has been manipulated, whereas an entirely black mask $(\M = 0)$ represents a pristine image.  

Our approach is to train a cGAN to create $\Mh$, an estimate of the forgery mask $\M$.  $\I$ is considered doctored if $\Mh \not \approx 0$, meaning that a forgery is detected in it and is comprised of the pixels located at $\{ (x,y) : \Mh(x,y) \neq 0\}$.  On the other hand, the image $\I$ is considered pristine if no forgery is detected, indicated by $\mathbf{\Mh \approx}$ 0.  Examples of satellite images $\I$ and their corresponding ground truth forgery masks $\M$ can be seen in Figure 1.

\begin{figure}[htbp]
    \centering
    \begin{minipage}{0.45\textwidth}
      \centering
      \subfloat[Pristine $\I$]{{\includegraphics[width=3cm]{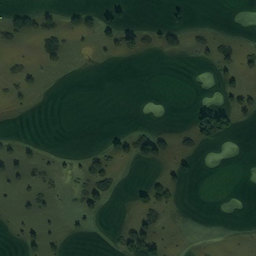} }}
      \qquad
      \subfloat[Pristine $\M$ ]{{\includegraphics[width=3cm]{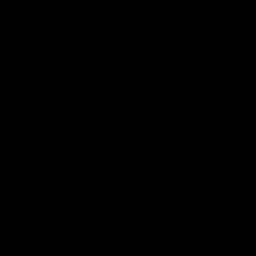} }}
    \end{minipage}\hfill
    \begin{minipage}{0.45\textwidth}
      \centering
      \subfloat[Forged $\I$]{{\includegraphics[width=3cm]{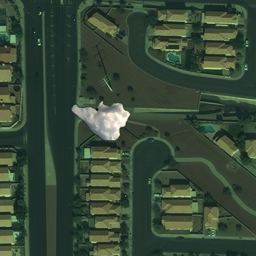} }}
    \qquad
    \subfloat[Forged $\M$ ]{{\includegraphics[width=3cm]{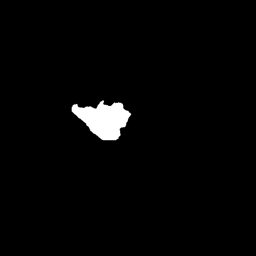} }}
    \end{minipage}
    \caption{Image - mask $\{ \I$, $\M \}$ pairs. (a) and (c) portray two example satellite images under analysis.  (b) and (d) illustrate their corresponding ground truth masks.  A pristine image's mask is entirely black, like (b).  (d) displays a mask that contains a forgery.  The cGAN will try to create mask estimates that resemble masks (b) and (d).}
\end{figure}

\section{Method}
In this section we describe our technique for splicing detection and localization. Additional details about the general cGAN concepts reported in this section can be found in \cite{pix2pix}. We train our cGAN on both pristine and forged images to learn a mapping from an input image $\I$ to a forgery mask $\M$.  It consists of two  parts: a generator $\textit{G}$ and a discriminator $\textit{D}$.  Figure 2 shows the overall cGAN architecture.

The generator $\textit{G}$ has a 16-layer U-net architecture (8 encoder layers, 8 decoder layers) with skip connections \cite{unet}.  When $\textit{G}$ is presented with an image $\I$, it computes an estimated forgery mask $\Mh$,  defined as $\Mh = \mathit{G}(\I)$. The generator's  objective  is to create  $\Mh$ that is close to the true $\M$. Meanwhile, the discriminator $\textit{D}$ is trained to differentiate between the true input-mask pairs $\{ \I$, $\M \}$ and synthesized input-mask pairs $\{ \I, \Mh \}$ coming from the generator. 
In a cGAN, the generator and the discriminator are coupled through a loss function. During the course of training the discriminator forces the generator to produce masks that are not only close to the ground truth but also good enough that the discriminator cannot distinguish them from the ground truth thus making the generator do a better job. 

The discriminator $\textit{D}$ has an architecture of a 5-layer CNN that does binary classification on masks.  
Sometimes, a true image-mask pair $\{ \I$, $\M \}$ is presented to $\textit{D}$.  Other times, an image-mask estimate pair $\{ \I, \Mh \}$ is presented.  In both cases, the image under analysis $\I$ is presented to the discriminator $\textit{D}$ along with either a true forgery mask $\M$ or a synthesized forgery mask $\Mh$.   $\textit{D}$ divides the input into patches of size 70x70 pixels.  It then classifies each patch as forged or pristine, assigning labels 0 and 1 respectively.  The values for all of the patches are averaged to determine the classification for the entire input.
The following equations describe the two cases outlined in this paragraph:\begin{equation} \label{eq2}
D(\mathbf{I}, \Mh) = D(\mathbf{I}, G\mathbf{(I)}) = 0,
\end{equation}\begin{equation} \label{eq3}
D(\mathbf{I}, \mathbf{M}) = 1.
\end{equation}
The generator $\textit{G}$ and the discriminator $\textit{D}$ compete in a min-max game, training and improving each other over time.  The coupled loss function of the network is described in the following equation:

\begin{equation} \label{eq4}
\begin{split}
\mathcal{L}_\text{cGAN}(G, D) = 
& \mathbb{E}_{\I,\M}[\log (D(\I, \M))] + \\
& \mathbb{E}_{\I,\text{\textbf{z}}}[\log(1 - D(\I, G(\I))].
\end{split}
\end{equation}

So far, we have described a network in which the generator $\textit{G}$ learns to create masks $\Mh$ that could be mistaken for real forgery masks by $\textit{D}$.  However, this does not ensure that the synthesized masks will correctly show forgeries in images. For example, $\Mh$ may ``fool'' $\textit{D}$  and be classified as an authentic mask for $\I$ without resembling its ground truth mask. In such a case, $\Mh \not \approx \M$. Therefore, we impose an additional constraint on the generator so that it learns to reconstruct the ground truth masks of training images, i.e., $\Mh \approx \M$ . This can be achieved by training $\textit{G}$ to minimize reconstruction loss $\mathcal{L}_\text{R}$ between $\Mh$ and $\M$. 
Since our task is to primarily classify every individual pixel into two classes (i.e., forged or pristine), we choose $\mathcal{L}_\text{R}$ to be a binary cross-entropy (BCE) loss term. This is different with respect to the classic pix2pix which uses $L_{1}$ as loss term $\mathcal{L}_\text{R}$. We later on verify in our experiments that BCE is indeed a better choice over $L_{1}$.
The total loss function of the cGAN is denoted as:\begin{equation} \label{eq5}
\mathcal{L} = \mathcal{L}_\text{cGAN} + \lambda \mathcal{L}_\text{R}.
\end{equation}
Once training is complete, the generator $\textit{G}$ is capable of producing masks $\Mh$ that are realistic and close to $\M$. To test new images under analysis, the discriminator is not considered, and the generator is used to produce mask estimates.

\begin{figure}[ht]
    \centering
    \begin{minipage}{0.45\textwidth}
      \centering
      \subfloat[Generator $\textit{G}$ coupled to discriminator $\textit{D}$ during training]{{\includegraphics[width=7cm]{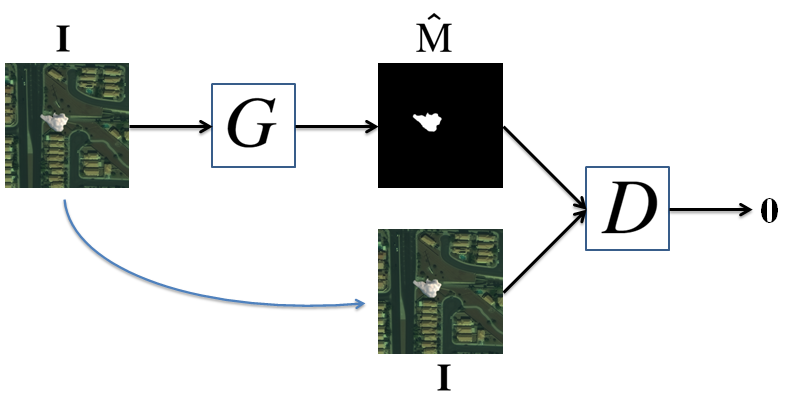} }}
    \end{minipage}\hfill
    \begin{minipage}{0.45\textwidth}
      \centering
      \subfloat[Discriminator $\textit{D}$ during training]{{\includegraphics[width=4cm]{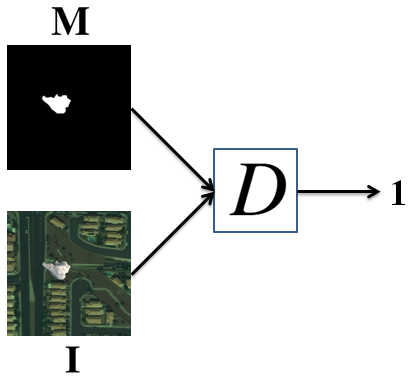} }}
    \end{minipage}
     \begin{minipage}{0.45\textwidth}
      \centering
      \subfloat[Generator $\textit{G}$ after training]{{\includegraphics[width=5cm]{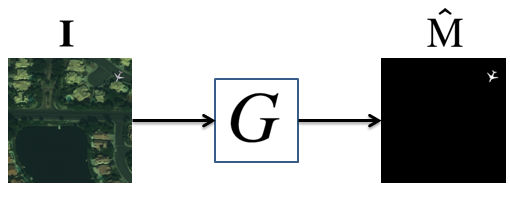} }}
    \end{minipage}
    \caption{cGAN architecture. (a) shows the $\textit{G}$-$\textit{D}$ training configuration, where $\textit{G}$ produces mask estimate $\Mh$ and presents it to $\textit{D}$ for evaluation.  $\textit{D}$ attempts to classify non-authentic $\{\I$, $\Mh\}$ pairs as 0.  (b) depicts $\textit{D}$ during training when presented with a true mask $\textbf{M}$.  It attempts to classify true $\{\I$, $\M\}$ pairs as 1.  The model resembles (c) after training is complete.}
\end{figure}

\begin{figure*}[ht]
    \centering
    
    \subfloat[Pristine $\I$]{{\includegraphics[width=3cm]{pristine_29_real_A} }} \qquad
    \subfloat[Small forgery $\I$]{{\includegraphics[width=3cm]{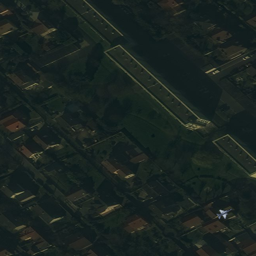} }} \qquad
    \subfloat[Medium forgery $\I$]{\includegraphics[width=3cm]{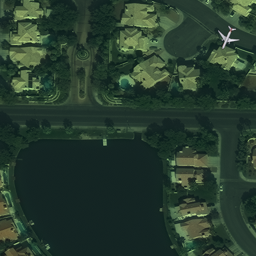}} \qquad
    \subfloat[Large forgery $\I$]{\includegraphics[width=3cm]{forged_138_128_real_A}}
    
    \subfloat[Pristine $\M$ ]{\includegraphics[width=3cm]{pristine_29_real_B}} \qquad
    \subfloat[Small forgery $\M$ ]{\includegraphics[width=3cm]{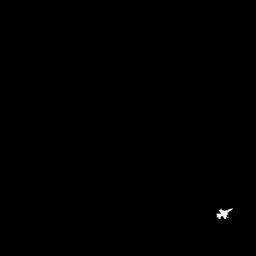}} \qquad
    \subfloat[Medium forgery $\M$ ]{\includegraphics[width=3cm]{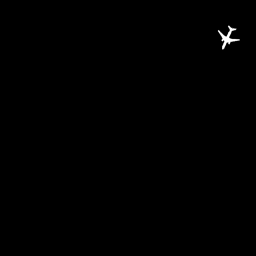}} \qquad
    \subfloat[Large forgery $\M$ ]{\includegraphics[width=3cm]{forged_138_128_real_B}} \qquad
    
    \subfloat[Pristine $\Mh$]{\includegraphics[width=3cm]{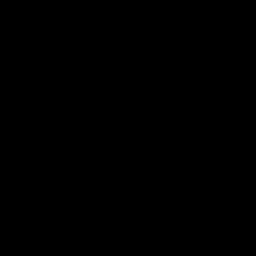}} \qquad
    \subfloat[Small forgery $\Mh$]{\includegraphics[width=3cm]{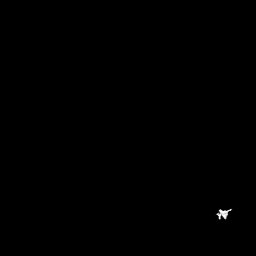}} \qquad
    \subfloat[Medium forgery $\Mh$]{\includegraphics[width=3cm]{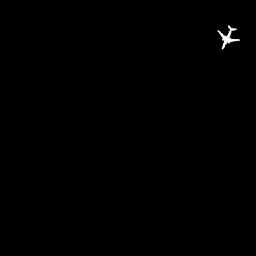}} \qquad
    \subfloat[Large forgery $\Mh$]{\includegraphics[width=3cm]{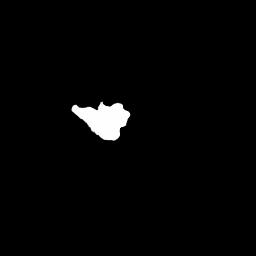}}
    
    \caption{Input images, ground truth masks, and generated mask estimates. Each column presents a set of $\{\I ,\M, \Mh\}$ example triplets for different types of input images from $\mathcal{D}_{P}$, $\mathcal{D}_{S}$, $\mathcal{D}_{M}$, $\mathcal{D}_{L}$, respectively.}
    \label{fig:examples}
\end{figure*}

\section{Experimental Validation}
In this section, we report the details of our experiments.  First, we describe the image dataset.  Next, training strategies are discussed.  Finally, we present experimental results and analysis.

We utilized the dataset presented in \cite{Sri-ei} for our experiments.  It contains color images of overhead scenes from a satellite and their corresponding ground truth forgery masks.  Each image-mask pair is defined as $\{ \I$, $\M \}$ and has resolution $650 \times 650$ pixels.  The images were adapted from ones originally provided by the Landsat Science program \cite{landsat, landsat2} run jointly by NASA \cite{NASA} and US Geological Survey (USGS) \cite{USGS}.  
To create forged images, objects such as airplanes and clouds  were spliced into some of the images at random locations.  These doctored images fall into one of three size categories (small, medium, or large) based on the approximate dimensions of the forgery they contain relative to the patch dimensions (70x70 pixels) used by the discriminator $\textit{D}$ to analyze a mask.  Small forgeries are approximately 32x32 pixels; medium forgeries are approximately 64x64 pixels, and large forgeries are approximately 128x128 pixels.  The remaining satellite images were left as pristine.  For our purposes, pristine and small-forgery samples underwent data augmentation to increase the size of the training dataset.  Augmentation methods included rotating pristine and small-forgery $\{ \I$, $\M \}$ pairs by multiples of 90$\degree$ and flipping them about the vertical and horizontal center axes.   This produced our dataset $\mathcal{D}$, which contains 344 total $\{ \I$, $\M \}$ pairs.  Also, 158 pairs contain small forgeries, 32 pairs contain medium forgeries, 31 pairs contain large forgeries, and 123 are pristine.  These subsets of $\mathcal{D}$ are denoted as $\mathcal{D}_{S}$, $\mathcal{D}_{M}$, $\mathcal{D}_{L}$, and $\mathcal{D}_{P}$, respectively.  Examples are shown in Figure 3.  

The dataset $\mathcal{D}$ was split into three sets for training, validation, and testing.  The training dataset $\mathcal{D}_{train}$ contains 128 $\mathcal{D}_{S}$ pairs and 90 $\mathcal{D}_{P}$ pairs.  The validation set $\mathcal{D}_{validation}$ has 32 $\mathcal{D}_{S}$ pairs and 18 $\mathcal{D}_{P}$ pairs.  The final dataset, $\mathcal{D}_{test}$, consists of 32 $\mathcal{D}_{M}$, 31 $\mathcal{D}_{L}$, and 15 $\mathcal{D}_{P}$ pairs.  By creating disjoint training/validation and evaluation datasets,  we observe how well a trained model extends to new forgery sizes.  It was hypothesized that small forgeries might pose the biggest challenge to the network, so they compose the training and validation sets. The cGAN was trained for 200 epochs using the Adam optimizer with an initial learning rate of 0.0002.  The reconstruction loss $\mathcal{L}_{R}$ coefficient $\lambda$ was set to 100.  After training, the model that performed the best on $\mathcal{D}_{validation}$ was selected to use for testing.

We did both visual and numerical analysis of the results to determine the effectiveness of our proposed method.  Figure~\ref{fig:examples} contains examples of mask estimates $\Mh$ produced by $\textit{G}$ and their corresponding ground truth masks $\M$. It shows that the model produces mask estimates of  both pristine and forged images that very closely resemble the ground truth masks, i.e., $\Mh \approx \M$.  Thus, we can clearly see if a forgery is present in an image $\I$ and, if so, its various properties.  A numerical analysis of the results further verifies this.  

\begin{figure}[ht]
    \centering
     \subfloat[ROC curves for forgery detection depicting comparison between BCE and $L_{1}$ loss for $\mathcal{L}_{R}$]{\includegraphics[width=\linewidth]{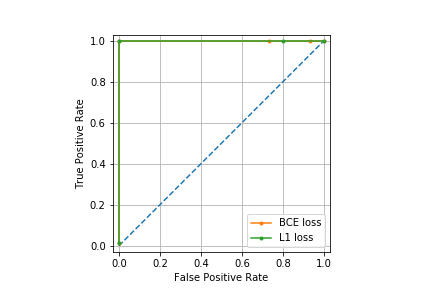}}
     
      \subfloat[ROC curves for forgery localization depicting comparison between BCE and $L_{1}$ loss for $\mathcal{L}_{R}$]{\includegraphics[width=\linewidth]{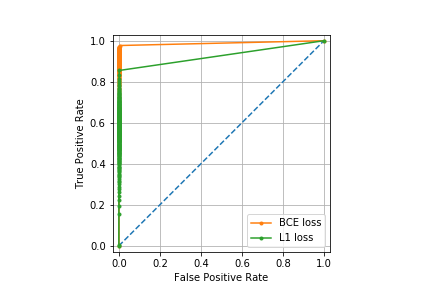}}
    \caption{ROC curves for detection and localization of spliced forgeries}
\end{figure}

\begin{figure}[ht]
\centering
\subfloat[PR curve for forgery detection using BCE loss for $\mathcal{L}_{R}$]{\includegraphics[width=\linewidth]{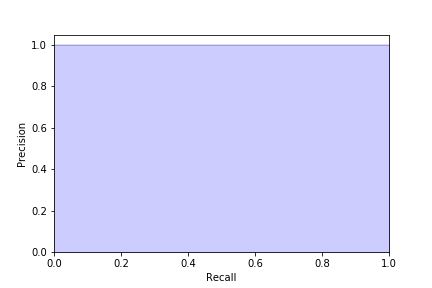}}

\subfloat[PR curve for forgery localization using BCE loss for $\mathcal{L}_{R}$]{\includegraphics[width=\linewidth]{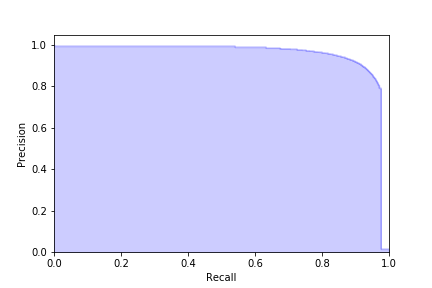}}
\caption{PR curves for detection and localization of spliced forgeries}
\end{figure}

To evaluate forgery detection, the average pixel value of a mask estimate is defined as
\begin{equation}
    \Mh_{avg} = \frac{1}{X\cdot Y} \sum_{x=1}^X \sum_{y=1}^Y \Mh(x, y),
\end{equation}
where $X \times Y$ is the image resolution.
Then, binary thresholding with threshold $\textit{T}$ is used to determine whether the image under analysis $\I$ is pristine or forged.  As  described above, an image is considered pristine when $\Mh \approx 0$.  From a thresholding standpoint, this is achieved when $\Mh_{avg} < \textit{T}$.  Otherwise, $\I$ is labeled as forged.  Figure 4 shows the receiver operating characteristic (ROC) curves that reveal the performance of different thresholds $\textit{T}$.  It also illustrates model performances achieved when using BCE loss and $L_{1}$ loss for reconstruction.  The areas under the curve (AUC)  for both BCE and $L_{1}$ loss are 1.000, indicating that it is possible to achieve perfect detection accuracy with thresholding.  These results are further verified by the precision-recall (PR) plot in Figure 5 for a model using BCE loss.  It too indicates that perfect detection is possible with our 2-class model, as its average precision score is also 1.000.

To assess forgery localization, a similar evaluation process occurs; however, only for images in which forgeries are detected.  Their mask estimates $\Mh$ are thresholded and then undergo a pixel-wise comparison to to their corresponding ground truth masks $\M$.  Figure 4 also shows ROC curves for localization  for different thresholds.  In this case, a performance difference in BCE $\mathcal{L}_{R}$ versus $L_{1}$ $\mathcal{L}_{R}$ is observed.  BCE yields a higher AUC value of 0.988 in comparison to $L_{1}$, which achieves an AUC of 0.927.  The PR curve (again using BCE loss) with an average precision score of 0.953 confirms that localization results are very good.  

\section{Conclusions}
In this paper, we propose a forensic image analysis method based on a cGAN for splicing detection and localization in satellite images. The proposed technique exploits a data driven approach, thus learns how to distinguish forged regions from pristine ones directly from the available training data.

Results show that the developed methodology accomplishes both tampering detection and localization with incredibly high accuracy on the used dataset. Moreover, it is interesting to notice how the proposed solution is able to generalize to forgeries of different sizes than those seen during training.

While the results of this experiment are very good, it would be interesting to see how the technique performs on different types of forgeries, as well as on datasets containing images coming from different satellites, to further test the method generalization capability.

\bibliographystyle{latex8}
\bibliography{latex8}

\end{document}